\documentclass[10pt,journal,compsoc]{IEEEtran}
\ifCLASSOPTIONcompsoc
  \usepackage[nocompress]{cite}
\else
  \usepackage{cite}
\fi

\ifCLASSINFOpdf
   \usepackage[pdftex]{graphicx}
   \DeclareGraphicsExtensions{.pdf,.jpeg,.png}
\else
\fi

\usepackage{times}

\usepackage{latexsym}
\usepackage{multirow}
\usepackage{booktabs}       % professional-quality tables
\usepackage{amsfonts}       % blackboard math symbols
\usepackage{amsmath}
\usepackage{wasysym}
\usepackage{algorithm}
\usepackage{algpseudocode}
\usepackage{nicefrac}       % compact symbols for 1/2, etc.
\usepackage{microtype}      % microtypography
\usepackage{array}
\usepackage{hyperref}
\usepackage{url}

\begin{document}
\title{EEG-based Emotional Video Classification via Learning Connectivity Structure}

\author{Soobeom Jang, Seong-Eun Moon,
  and~Jong-Seok Lee,~\IEEEmembership{Senior Member,~IEEE}% <-this % stops a space
  \IEEEcompsocitemizethanks{\IEEEcompsocthanksitem The authors are with the School of Integrated Technology, Yonsei University, Korea.\protect\\
    E-mail: \{soobeom.jang, se.moon, jong-seok.lee\}@yonsei.ac.kr
  }% <-this % stops an unwanted space
  \thanks{Manuscript received April 19, 2005; revised August 26, 2015.}}


\markboth{Journal of \LaTeX\ Class Files,~Vol.~14, No.~8, August~2015}%
{Shell \MakeLowercase{\textit{et al.}}: Bare Demo of IEEEtran.cls for Computer Society Journals}



\IEEEtitleabstractindextext{%
  \begin{abstract}
    Electroencephalography (EEG) is a useful way to implicitly monitor the user's perceptual state during multimedia consumption.
    One of the primary challenges for the practical use of EEG-based monitoring is to achieve a satisfactory level of accuracy in EEG classification.
    Connectivity between different brain regions is an important property for the classification of EEG.
    However, how to define the connectivity structure for a given task is still an open problem, because there is no ground truth about how the connectivity structure should be in order to maximize the classification performance.
    In this paper, we propose an end-to-end neural network model for EEG-based emotional video classification, which can extract an appropriate multi-layer graph structure and signal features directly from a set of raw EEG signals and perform classification using them.
    Experimental results demonstrate that our method yields improved performance in comparison to the existing approaches where manually defined connectivity structures and signal features are used.
    Furthermore, we show that the graph structure extraction process is reliable in terms of consistency, and the learned graph structures make much sense in the viewpoint of emotional perception occurring in the brain.
  \end{abstract}

  }

\maketitle

\IEEEdisplaynontitleabstractindextext

% \fi
\IEEEpeerreviewmaketitle

\IEEEraisesectionheading{\section{Introduction}\label{sec:introduction}}

\IEEEPARstart{A}{nalyzing} brain activity through brain imaging such as electroencephalography (EEG) is important in understanding the affective mental state, such as emotion or thought of humans.
It is essential in a variety of applications including brain-computer interface, emotion recognition, and mental disease diagnosis.
In particular, implicit monitoring using EEG during multimedia consumption has received much attention in recent days for its potential applications such as real-time multimedia content recommendation and filtering \cite{moon2016implicit}.

The brain consists of multiple functional regions, and the activation patterns over the regions provide valuable information regarding the mental state.
Therefore, studying the inter-regional relationship appearing in the patterns, which is called functional connectivity (simply noted as connectivity in this paper), has been shown to be effective for analysis of the brain signal \cite{bullmore2011brain,horwitz2003elusive}.
Because the physical closeness of brain regions does not guarantee high connectivity between them, the connectivity structure needs to be analyzed in domains other than the Euclidean space.
Thus, a graph is the most natural and suitable data structure to represent the connectivity.
Previous studies showed that the graph analysis approach can be successfully applied to understanding of brain signals \cite{betzel2017multi,preti2017dynamic,rubinov2010complex}.

However, how to measure the level of connectivity, how to define an appropriate graph structure, and how to define appropriate signal features from different brain regions are all still open problems.
Usually, they are determined manually based on a priori knowledge.
For instance, correlation or causality metrics between the signals from different brain regions can be used as connectivity measures \cite{ding2006granger}. Then, a graph can be constructed by connecting pairs of the regions showing large connectivity values \cite{bullmore2011brain}, and finally, the power or entropy of the signal can be used as a feature of each region (i.e., a vertex of the graph) \cite{saa2010eeg,sabeti2009entropy}.
% Apparently, however, separately tackling these issues manually would not be optimal.
Apparently, however, it would not be optimal to separately tackle these issues manually.
In fact, this challenge is applied to not only brain signal data but also other data involving graph structures such as social networks and chemicals \cite{battaglia2018relational,de2018molgan,franceschi2019learning}.

In this paper, we resolve these issues via learning from data.
We propose a new deep learning model, which can extract both a graph structure and a feature vector at each vertex of the graph from raw time-series EEG data, and perform classification using the extracted graph and features.
The whole model is trained in an end-to-end manner to maximize the classification performance, which can simultaneously optimize the graph extraction module, the feature extraction module, and the classification module using a graph neural network.
The graph and features become all different adaptively for different data instances, which can overcome the notorious non-task-related variability of EEG signals.
In particular, the extracted graphs are weighted directed multi-layer graphs, which can convey rich information regarding the brain activation pattern better than undirected single-layer graphs.

% \color{blue}
There have been previous deep learning-based attempts for learning connectivity structures for EEG classification \cite{song2018eeg,wang2018eeg,zhang2019gcb,zhong2020eeg}.
However, they are limited in that only one graph structure is learned for all data and thus sample-specific structures are not modeled, manually determined initial graph structures need to be provided, only single-layer graphs are learned, and hand-crafted signal features need to be provided.
The work in \cite{song2020instance} proposes a method to learn multi-layer sample-specific structures, but the number of layers is restricted to the dimension of hand-crafted signal features.
To the best of our knowledge, our proposed method is the first to jointly learn sample-specific multi-layer connectivity structures and signal features directly from raw EEG data without the necessity of a prior knowledge regarding the graph structures and features.
% \color{black}

In summary, the main contributions of this work are as follows:

\begin{itemize}
  \item We propose a novel end-to-end neural network model for classifying brain signals, which learns both the connectivity structure and signal features.
        For learning the connectivity structure, we introduce three types of graph sampling methods to control the restriction in the extracted graph structures.
        To extract useful signal features from the raw time-series EEG data, we propose a 1-D dilated inception block that models the traditional frequency band-based brain signal processing technique.
  \item We suggest a way to evaluate the quality of the extracted graph structure in terms of consistency, in order to alleviate the limitation that no ground truth structure is available to measure the correctness of the obtained graph structure.
  \item We present several perspectives to analyze extracted graph structures, including graph consistency for quantitative analysis and neuroscientific analysis for qualitative analysis.
\end{itemize}

The rest of the paper is organized as follows. Section \ref{sec2:relatedworks} presents the background and related work. Section \ref{sec3:method} describes the proposed method, which includes graph membership extraction, graph sampling, feature extraction, and graph neural network. Section \ref{sec4:experiments} shows our experiment to examine the performance of the proposed method and evaluates the extracted connectivity structures through quantitative and qualitative analyses. Finally, Section \ref{sec5:conclusion} concludes the paper.

\section{Related Work}
\label{sec2:relatedworks}

\subsection{Deep learning for brain signal} There are two major approaches to brain signal analysis and classification using deep learning models: signal-based approach and connectivity-based approach.
The signal-based approach exploits popular convolutional neural networks (CNNs) and recurrent neural networks (RNNs), to which the time-series brain signal data (themselves or after conversion to image-like representations) are inputted, without considering inter-regional connectivity.

%%%%%%%%%%%%%%%%%%%%%%%%%%%%%%%%%%%%%%%%%%%%%%%%%%%%%%%%%%%%%%%%

Some studies use raw signals directly to extract time-domain features using CNNs.
In \cite{schirrmeister2017deep}, a CNN architecture using both 1-D and 2-D convolutional layers is proposed for motor imagery classification.
The model uses a 1-D convolutional layer to extract low-level time-domain features, while a 2-D convolutional layer is adopted to integrate information from multiple electrodes.
The study in \cite{lawhern2018eegnet} adopts a similar approach.
1-D and 2-D convolutional layers are used to extract the time-domain feature and its spatial integration, respectively, and additional separable 1-D convolution is applied to learn temporal filters for each spatial filter.
The proposed CNN architecture is applied to brain-computer interface problems such as classification of event-related potentials and motor imagery.
% \color{blue}
In \cite{cui2020eeg}, a CNN architecture for emotion classification is proposed, called the regional-asymmetric convolutional neural network (RACNN), which includes an asymmetric feature extractor to model differences between the two hemispheres.
The model maps electrodes to pixels of a 3-D tensor, and applies convolutional layers to extract temporal features.
For classification, asymmetric features and spatial features are computed based on the extracted temporal features.
% \color{black}

RNNs including long short-term memory (LSTM) networks are another natural choice to model the temporal characteristics of raw EEG signals.
% In \cite{alhagry2017emotion}, a two layer LSTM network is proposed and is used for emotion recognition.
For example, the work in \cite{bashivan2015learning} proposes a recurrent convolutional neural network.
The network uses a CNN that processes spectral topography maps for each time frame.
The outputs of multiple frames are fed into an LSTM network to obtain features considering both the spectral-domain and time-domain.
In \cite{zhang2018cascade}, a similar architecture is proposed, which uses spatial topography maps to construct the time-series maps for the input.
% \color{blue}
The study in \cite{li2019regional} proposes a bidirectional LSTM-based method to model regional dependence and temporal relationships for emotion recognition.
To compute regional dependence, the electrodes are grouped into multiple regions for modeling regional features.
Then, an attention mechanism over regions is applied to compute the importance of each region, which is concatenated with temporal features.
% \color{black}

%%%%%%%%%%%%%%%%%%%%%%%%%%%%%%%%%%%%%%%%%%%%%%%%%%%%%%%%%%%%%%%%

Noting that the brain connectivity provides valuable information in understanding the brain activity, recent studies try to extract connectivity patterns from EEG and use them as features for classification.
In some studies, the connectivity information representing pairwise correlation or causality is represented as images, which are modeled by CNNs.
The work in \cite{moon2018convolutional} evaluates various connectivity measures, including Pearson correlation, phase-locking value, and phase lag index, using 2-D CNNs for emotion recognition.
The work in \cite{phang2019classification} proposes a 2-D CNN to detect schizophrenia using vector autoregressive model-based connectivity features and graph metrics such as the clustering coefficient.

\subsection{Graph neural networks (GNNs)}

Graph neural networks (GNNs) are a type of neural network to process graph data that are not well defined in the Euclidean space \cite{wu2019comprehensive}.
The main mechanism of graph neural networks is to integrate the information of a target vertex with that of neighboring vertices, which is modeled as the message passing operation \cite{gilmer2017neural}.
This can be seen as a graph domain adaptation of the convolutional operation performed on neighboring pixels in CNNs \cite{kipf2016semi,defferrard2016convolutional}.
GNNs are widely used in various applications involving non-Euclidean data, including analysis of chemicals \cite{coley2019graph}, citation networks \cite{velivckovic2017graph}, recommender systems \cite{wang2019neural}, and skeleton-based action recognition \cite{yan2018spatial}.

% \color{blue}
Recently, a few methods have been proposed to obtain connectivity-based graph representations of brain signals and model them using GNNs.
In order to apply a GNN to EEG, a prerequisite is to define an appropriate graph structure consisting of vertices and edges.
Hence, how to define a graph structure and vertex features is a main research problem of the previous work.
The work in \cite{jang2018eeg} presents several methods to define vertex features and graph structures from EEG data, for which a GNN called ChebNet is used to perform EEG-based video identification.
The signal power or entropy is used as the vertex feature, and the graphs whose edge weights are determined using the signal correlation or distance between electrodes are considered to obtain the graph structure.
In \cite{song2018eeg}, a method called the dynamic graph CNN (DGCNN) is proposed for emotion recognition, which is a GNN optimizing both the GNN model and graph structure.
The graph structure is initialized using a k-nearest neighbor graph based on the Euclidean distance between electrodes and is modified through the learning process.
In \cite{wang2018eeg} and \cite{zhang2019gcb}, GNNs are used to extract features from EEG, and the so-called broad learning system is used for classification.
% \color{black}

\subsection{Graph structure inference}
Inferring a graph structure from the given non-graph data is important to understand the nature of the data in many applications \cite{brugere2018network}, but is still an open problem \cite{battaglia2018relational,qiao2018data}.
Traditionally, this has been dealt with using probabilistic models or statistical inferences \cite{belilovsky2017learning,mei2016signal,varoquaux2010brain}.

% \color{blue}
Recently, neural network-based methods have been proposed.
The neural relational inference \cite{kipf2018neural} is a variational autoencoder-based model to extract structural information of dynamic relational systems such as physical interaction systems for prediction of future states of dynamic objects from observed past states.
However, this method requires a priori knowledge about the graph structure (i.e., graph density), whereas our method relies only on the data to extract an appropriate graph structure.
In \cite{franceschi2019learning}, a graph convolutional network-based model for transductive learning is proposed, which jointly learns the graph structure and features using bilevel programming for vertex classification problems.

For EEG, a few methods for graph structure inference have been proposed.
The DGCNN method explained above is one such example, which requires a manually designed initial graph structure and hand-crafted features.
The work in \cite{zhong2020eeg} proposes a similar approach with asymmetricity modeling in the initial graph structure.
In these methods, a single-layer graph structure, which expresses connectivity of only a single aspect, is estimated for all data, thus it is difficult to incorporate sample-specific characteristics in classification.
The work in \cite{song2020instance} proposes an instance-adaptive graph (IAG) method, which allows to obtain sample-specific multi-layer graph structures.
However, the hand-crafted features need to be extracted to obtain the graphs, and the number of graph layers is restricted by the feature dimension (i.e., frequency bands).
Different from these existing studies, our proposed method jointly learns the multi-layer graph structure and features specific to each EEG sample from scratch without any a prior knowledge about the graph structure so that multiple connectivity relationships in EEG can be simultaneously modeled.
% Furthermore, while the existing methods learn single-layer graphs that can express connectivity of only a single aspect, our method produces multi-layer graphs so that multiple connectivity relationships in EEG can be simultaneously modeled. % TODO
% \color{black}

\section{Proposed Method}
\label{sec3:method}

\subsection{Problem statement}

Our goal is to build a neural network model performing the classification of the given EEG data by discovering the underlying connectivity structure. Since no ground truth or hint about the underlying connectivity structure is provided, the model needs to estimate it from the data.
In particular, the connectivity structure learned by our method has the following characteristics.
First, the connectivity structure is estimated for each observation (data instance) separately in order to accommodate the fact that different connectivity structures result in different activation patterns over the brain.
Second, we obtain multiple types of connectivity involved in the brain activation from each observation.
Mathematically, the estimated connectivity structure has the form of a multi-layer graph instead of a single-layer graph.
Here, the layers share common vertices corresponding to EEG electrodes, but each layer consists of a unique set of edges, which corresponds to a separate type of connectivity.

The given observation is represented by $(X,y)$, where $X \in \mathbb{R}^{N\times T}$ is a set of time-domain signals collected from $N$ electrodes, whose length is $T$, and $y$ is the corresponding class label.
We assume that the graph structure to be estimated is a weighted directed multi-layer graph without self-loops:
\begin{equation}
  \phantom{.} G=(V, \{E^k\}_{k=1}^K, \{W^k\}_{k=1}^K) .
\end{equation}
Here, $V=\{v_i\}_{i=1}^N$ represents the vertices, and $E^k \in \{0,1\}^{N^2}$ and $W^k=\{w_{ij}^k\}_{i,j=1}^N \in \mathbb{R}^{N\times N}$ indicate the existence of edges and the edge weights between vertex pairs in the $k$th graph layer, respectively.
We assume $0 \leq w_{ij}^k \leq 1$. $K$ is a hyperparameter to control the number of graph layers.
Once the graph structure is estimated, it is used together with features separately extracted from $X$ in order to predict $y$.

\subsection{Proposed model}

\begin{figure*}[!t]
  \small
  \centering
  \includegraphics[width=\textwidth]{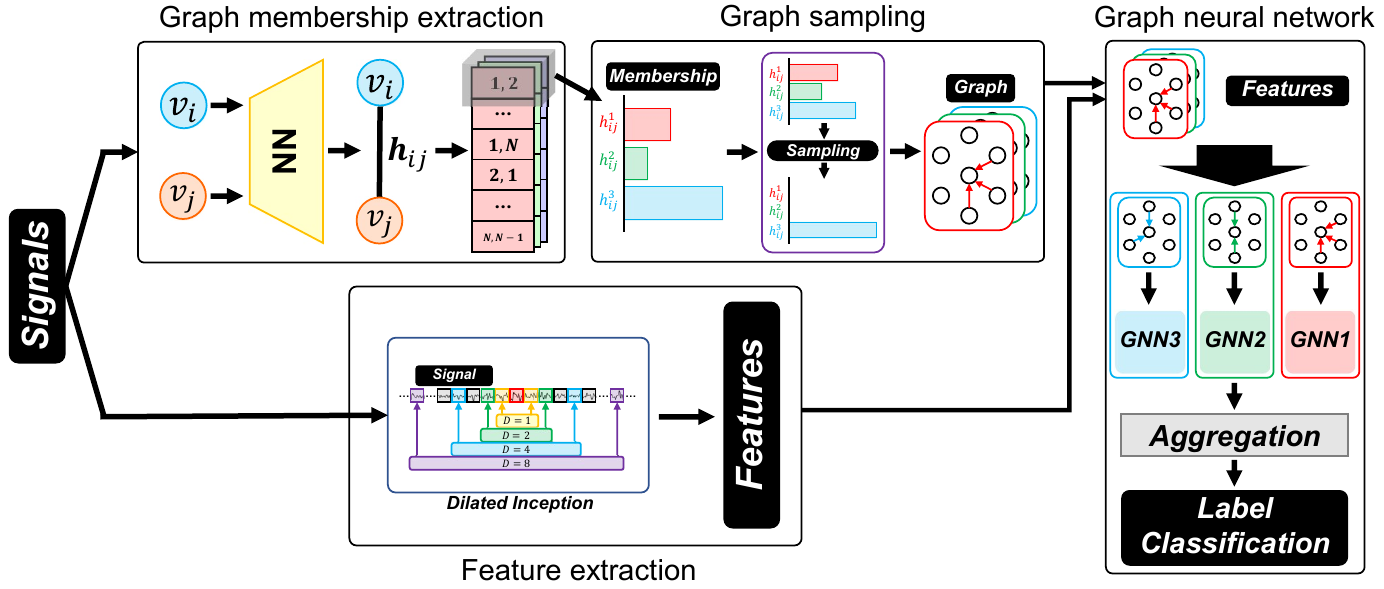}
  \caption{\label{fig:overall} Overall structure of the proposed model.}
\end{figure*}

The overall structure of our model is illustrated in Figure \ref{fig:overall}. It can be divided into four modules: graph membership extraction, graph sampling, feature extraction, and classification.

In the graph membership extraction module, the given set of raw EEG signals are fed into a neural network to infer the membership of each edge to each connectivity type (i.e., graph layer).
The inferred connectivity memberships are processed by the graph sampling module to construct a multi-layer graph structure.
In the feature extraction module, the raw EEG signals are converted to features using convolutional and max-pooling operations.
Finally, the obtained graph structure and features are combined in a GNN to obtain the final classification result.

\subsubsection{Graph membership extraction}

From the given EEG signals, we compute the latent membership $\mathbf{h}_{ij}=[h_{ij}^1, \ldots, h_{ij}^K]$ representing the certainty of the existence of connectivity from vertex $v_i$ to vertex $v_j$ for each graph layer (i.e., connectivity type).
Let $\mathbf{x}_{i}, \mathbf{x}_{j} \in \mathbb{R}^{T}$ be the signals recorded from electrodes $i$ and $j$, which are the input vectors of vertices $v_i$ and $v_j$, respectively.
First, a feature vector is extracted from each of them by a shared fully-connected neural network ($f_1$), which produce $f_1(\mathbf{x}_{i})$ and $f_1(\mathbf{x}_{j})$, respectively.
Then, the features are processed by another fully-connected neural network ($f_2$) after concatenation, i.e.,
\begin{equation}
  \label{eq:me_n2e_f2}
  \phantom{,} \mathbf{r}_{ij} = f_2 \big(  f_1(\mathbf{x}_i) \| f_1(\mathbf{x}_j) \big) ,
\end{equation}
where $\|$ denotes the concatenation operation.
The result $\mathbf{r}_{ij}$ can be seen as an edge feature for the potential edge between $v_i$ and $v_j$.
We aggregate this edge feature vector between vertex $v_i$ and the other vertices and perform a nonlinear transform to obtain the vertex feature for $v_i$:
\begin{equation}
  \label{eq:me_e2n_f3}
  \phantom{,} \mathbf{h}_{i} = f_3 \left( \sum_{\substack{j=1 \\ j \neq i}}^{N} \mathbf{r}_{ij} \right) ,
\end{equation}
where $f_3$ is a fully-connected neural network.
Note that the above steps compose a message passing operation \cite{gilmer2017neural}.
Finally, the latent connectivity membership from $v_i$ to $v_j$ is obtained by a nonlinear transform (using a fully-connected network) of the vertex features of $v_i$ and $v_j$ and the edge feature:
\begin{equation}
  \label{eq:me_n2e_f4}
  \phantom{,} \mathbf{h}_{ij} = f_4 \big( \mathbf{h}_{i} \| \mathbf{h}_{j} \| \mathbf{r}_{ij} \big) ,
\end{equation}
where $f_4$ is a fully-connected neural network.

\subsubsection{Graph sampling}
\label{sec:graph_sampling}

From the graph layer membership information, we construct a multi-layer graph structure via the graph sampling module.
We consider the following three methods for sampling: stochastic sampling, deterministic thresholding, and continuous softmax.
One of the three methods is applied to the graph layer membership information $\mathbf{h}_{ij}$ to obtain a graph structure.

\begin{figure}[!b]
  \small
  \centering
  \includegraphics[width=0.99\columnwidth]{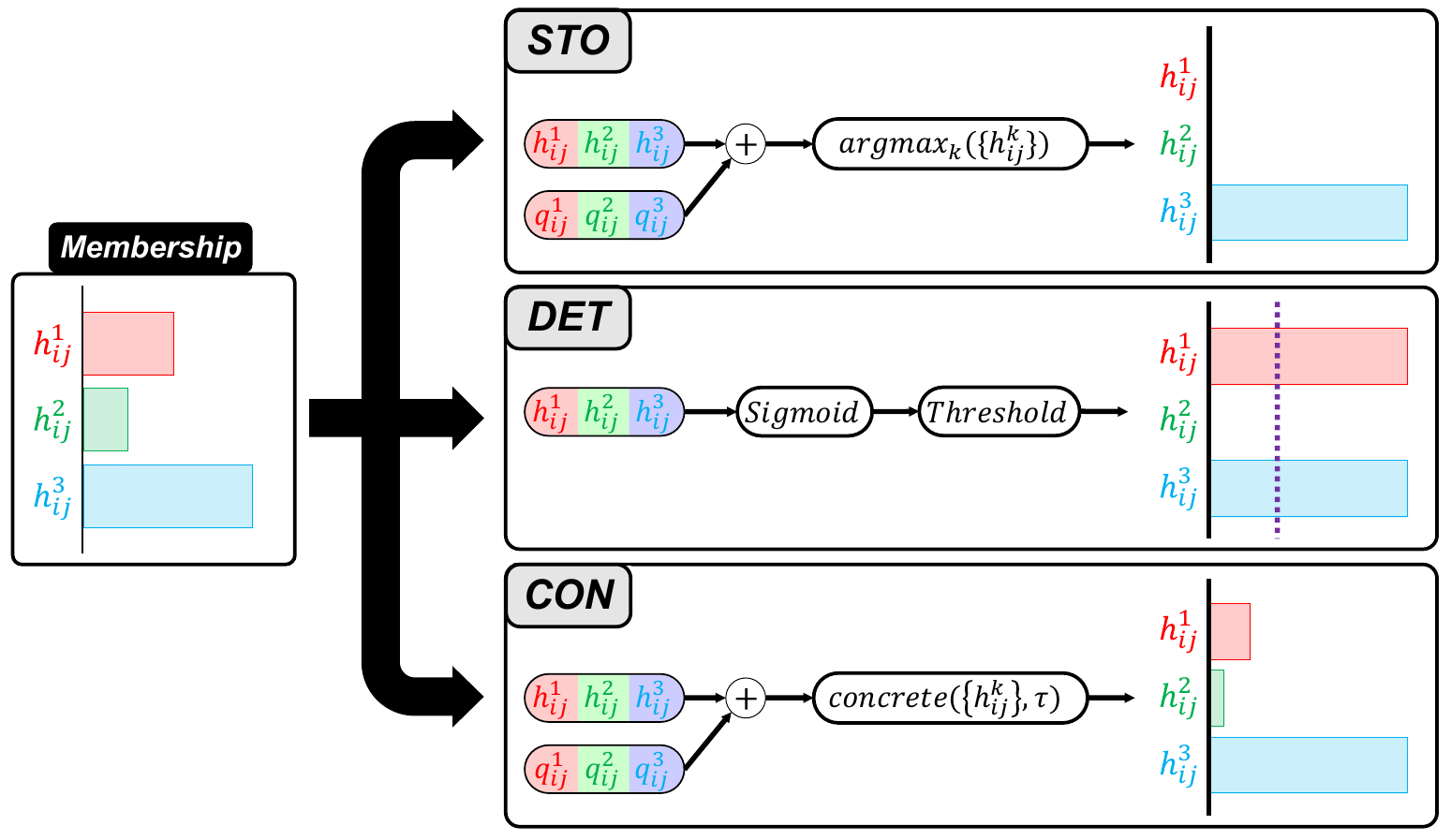}
  \caption{\label{fig:graph_sampling} Three graph sampling methods. Note that in the stochastic sampling method (STO), the division by $\tau$ and the softmax operation in (\ref{eq:gumbel}) are not required since they do not affect the subsequent argmax operation.}
\end{figure}

A schematic diagram of the three methods is shown in Figure \ref{fig:graph_sampling}.
The main difference among them is the degree of restriction in activating multiple graph layers, which is detailed below.

\noindent
\textbf{Stochastic sampling (STO).} The stochastic sampling method probabilistically assigns the potential edge from vertex $v_i$ to vertex $v_j$ to one of the $K$ graph layers.
In this case, the sampled graph weight is binary (0 or 1), which is not differentiable directly.
Therefore, the Gumbel-softmax reparametrization technique \cite{jang2016categorical,maddison2016concrete} is used to provide continuous relaxation and enable computation of gradients:
\begin{equation}
  \phantom{,} \mathbf{z}_{ij}= \mathrm{softmax}\left( \frac{\mathbf{h}_{ij}+\mathbf{q}}{\tau} \right) ,
  \label{eq:gumbel}
\end{equation}
where $\mathbf{q}\in \mathbb{R}^K$ is a random vector whose components are i.i.d. and follow the standard Gumbel distribution.\footnote{We observed that the obtained graph structure and the corresponding classification result rarely change due to the sampling during inference ($<$1\% for graph structure and $<$0.2\% for accuracy across 10 repetitions).}
$\tau$ is the softmax temperature controlling sampling smoothness, which is set as $\tau=0.5$ in this paper as a reasonable choice to achieve both a good approximation of categorical distribution and stable learning \cite{jang2016categorical,maddison2016concrete}.
Then, the unweighted edge from $v_i$ to $v_j$ is obtained by choosing the maximum value among the components of $\mathbf{z}_{ij}$, i.e.,
\begin{equation}
  \phantom{,} w_{ij}^k= \left\{ \begin{array}{ll}
    1 & \mbox{if $k=\mathrm{argmax}_{l \in \{1, ..., K\}} (z_{ij}^l)$} \\
    0 & \mbox{otherwise}\end{array} , \right.
  \label{eq:stochastic_binarization}
\end{equation}
where $z_{ij}^{l}$ is the $l$th component of $\mathbf{z}_{ij}$.
The stochastic sampling method is the most restrictive to construct a graph structure, enforcing only one of the graph layers to be activated.
That is, it selects the most likely connectivity type for each edge candidate.

\noindent
\textbf{Deterministic thresholding (DET).} The estimated graph has multiple layers, and we expect that different graph layers model different types of connectivity information.
For some vertex pairs, their relationship may be a mixture of multiple connectivity types, which requires edges in multiple graph layers for representation.
The deterministic thresholding method relaxes the restriction of stochastic sampling to allow a pair of vertices to have edges in multiple layers via thresholding:
\begin{equation}
  \phantom{,} w_{ij}^k= \left\{ \begin{array}{ll}
    1 & \mbox{if sigmoid$(h_{ij}^k) > \Delta$} \\
    0 & \mbox{otherwise}\end{array} , \right.
\end{equation}
where $\Delta$ is a threshold, which is set as $\Delta=0.5$ in our work.
The same continuous relaxation technique used in the stochastic sampling method is used to make discrete variables differentiable during training.

In the stochastic sampling method, one of the graph layers must be activated even though any of the memberships are not significant.
On the other hand, deterministic sampling allows not only the activation of multiple graph layers for a single edge candidate but also the deactivation of all graph layers if all the membership values are insignificant.

\noindent
\textbf{Continuous softmax (CON).} The continuous softmax is the least restrictive method, which allows continuous edge weight values for all graph layers.
For this, $z_{ij}^k$ $(k=1, ..., K)$ having a continuous value between 0 and 1, which is obtained from the Gumbel-softmax operation in (\ref{eq:gumbel}), is directly used as the edge weight from $v_i$ to $v_j$ in the $k$th graph layer.
While the previous two methods construct unweighted graphs, the continuous softmax method generates weighted graphs so that distinct degrees of connectivity are maintained in different graph layers.
% \color{blue}
Note that this method is equivalent to the stochastic sampling method without the binarization process described in (\ref{eq:stochastic_binarization}).
% \color{black}
Therefore, this method produces the most general form of graph structures among the three methods, i.e., weighted directed multi-layer graphs.

\noindent
\textbf{Skip layer.} A graph constructed by the stochastic and continuous softmax methods enforces every pair of vertices to have edges in at least one graph layer.
However, there may exist no direct relationship between certain pairs of vertices.
In order to consider such cases, one of the graph layers can be assigned as a skip layer. The skip layer is discarded when the graph is passed to the GNN so that the edges belonging to this layer are omitted in the graph used for classification.

\subsubsection{Feature extraction}

\begin{figure}[!b]
  \small
  \centering
  \includegraphics[width=\columnwidth]{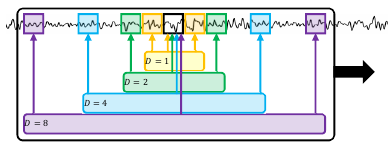}
  \caption{\label{fig:dilatedinception} Illustration of our 1-D dilated inception block. $D$ corresponds to the dilation rate.}
\end{figure}

The feature extraction module converts the raw EEG signals to features for end-to-end learning.
The main purpose of feature extraction is to capture the dynamic information in the signals.
Here, it is beneficial to extract features representing dynamic information at various temporal rates (from slow-varying to fast-varying factors), from which the GNN can selectively learn useful information for classification.
Different rates can be considered using time-domain windows having different lengths, or equivalently, different sizes of temporal receptive fields.
However, widely used CNN and RNN architectures typically rely on small-sized local receptive fields in each layer.
Thus, to examine slow-varying factors, a large number of layers is necessary, which is computationally inefficient \cite{wang2018non}.

In order to capture various dynamic characteristics from the signals, we adopt a 1-D version of the dilated inception block, which includes dilated convolutional layers with various dilation rates \cite{shi2017single,yang2019dilated,Yu2015multi}.
Our 1-D dilated inception block is illustrated in Figure \ref{fig:dilatedinception}.

The block consists of multiple 1-D convolutional layers having different dilation rates. A dilated convolutional layer with a low dilation rate captures features appearing among neighboring samples, which correspond to fast-changing high-frequency information, while that with a high dilation rate considers slow-varying features over a large temporal window.
This can be seen as analogous to the popular EEG signal analysis approach where the signal is divided into multiple frequency bands for separate frequency-selective analysis \cite{barry2009eeg,harmony1996eeg}.
The outputs of the convolutional layers are concatenated.
As a result, we obtain features for each vertex $U_i=[\mathbf{u}_{i1}, \ldots, \mathbf{u}_{iT'}] \in \mathbb{R}^{T'\times F}$ ($i=1,...,N$), where $T'$ and $F$ correspond to the time length and the number of features of the output, respectively.
The values of $T'$ and $F$ are determined by the structure of the feature extraction module; we describe our implementation in Section \ref{sec:model_details}.

\subsubsection{Graph neural network}

Finally, the GNN performs classification using the constructed graph structure and the extracted signal features by generating a vector $\hat{y}$ representing the class probabilities.
Since we have a multi-layer graph structure, the GNN is applied for each graph layer and the results are aggregated.
For the $k$th graph layer, the output of the GNN for vertex $v_i$ at time $t$, $\mathbf{s}^{k}_{it} \in \mathbb{R}^{F'}$, is obtained as follows.
\begin{equation}
  \label{eq:gnn_mp}
  \phantom{,} \mathbf{s}^{k}_{it} = \sum_{\substack{j=1 \\ j \neq i}}^{N} w_{ij}^k g_1^k( \mathbf{u}_{it}\| \mathbf{u}_{jt}) ,
\end{equation}
where $g_1^k$ is a graph layer-wise nonlinear transformation modeled by a fully-connected network having rectified linear units (ReLU) and $F'$ is a hyperparameter determining the feature dimension of the output.
In other words, a message passing operation is applied for vertex $v_i$ over the neighboring vertices by concatenating the features of each vertex pair ($\mathbf{u}_{it}$ and $\mathbf{u}_{jt}$) and aggregating them according to the edge weights ($w_{ij}^k$) after nonlinear transformation using $g_1^k$.
The outputs of the GNN for the $K$ graph layers are aggregated by addition:
\begin{equation}
  \label{eq:gnn_aggregation}
  \phantom{.} \mathbf{s}_{it} = \sum_{k=L}^K \mathbf{s}^{k}_{it} .
\end{equation}
Here, $L=2$ if a skip layer is used (assuming that the first graph layer is assigned as the skip layer) and $L=1$ otherwise.
At the final stage, a skip connection is used to ease the training process and encourage feature reuse as in DenseNet \cite{huang2017densely}.
In other words, the aggregated GNN outputs are concatenated with the signal features, which is then fed into a fully-connected network having the softmax activation function, $g_2$, i.e.,
\begin{equation}
  \phantom{,} \hat{y} = g_2 \Big( \mathrm{vec}\left(\{ U_i \}_{i=1}^N \right) \| \mathrm{vec}\left( \{S_i \}_{i=1}^N \right) \Big) ,
\end{equation}
where $S_i=[\mathbf{s}_{i1}, \ldots, \mathbf{s}_{iT'}]$, and $\mathrm{vec}()$ is the vectorization operation.

\section{Experiments}
\label{sec4:experiments}

% \color{blue}
This section describes our experiments for the EEG-based emotional video classification task defined in \cite{jang2018eeg}.
This task requires to learn comprehensive emotional responses specific to each video stimulus, which is more challenging than the traditional binary valence or arousal classification tasks.
The performance of the proposed model is compared with that of existing methods and is analyzed through ablation studies. Moreover, the connectivity structures learned by the proposed model are examined by graph consistency analysis and qualitative analysis.
% \color{black}

\subsection{Experimental setup}

\begin{figure}[!t]
  \small
  \centering
  \includegraphics[width=0.99\columnwidth]{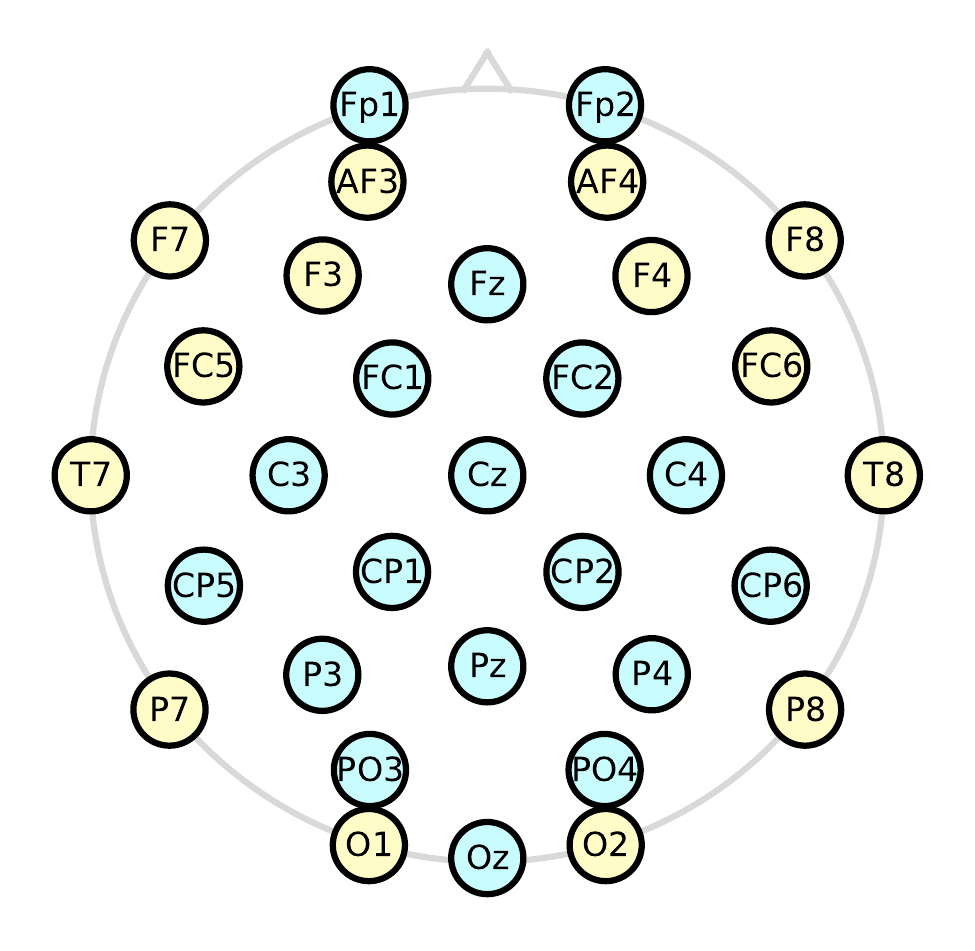}
  \caption{\label{fig:electrodes} Names and locations of the EEG electrodes. The yellow-colored vertices are used in both the DEAP database and the DREAMER database, and the blue-colored vertices are used only in the DEAP database.}
\end{figure}

\subsubsection{Databases}

We use two EEG databases regarding human affective mental states: DEAP \cite{koelstra2011deap} and DREAMER \cite{katsigiannis2017dreamer}.
The DEAP database is one of the largest databases for affective mental state analysis, containing 32-channel EEG recordings collected from 32 subjects during watching 40 affective video stimuli.
The DREAMER database includes EEG data collected using a 14-channel portable low-cost EEG device from 23 subjects for 18 video stimuli.
The signals are sampled at a rate of 128Hz for both databases.
% \color{blue}
Since we consider the video identification task where the model classifies the given set of EEG signals to one of the video stimuli used for mental state recording, the number of classes is 40 (for DEAP) or 18 (for DREAMER).
% \color{black}
Figure \ref{fig:electrodes} shows the names and locations of the electrodes for the two databases.

% \color{blue}
We use the first sixty-second-long signal from each EEG signal.
We construct five-fold cross-validation experiments by dividing each signal into five temporal segments (0-12s, 12-24s, 24-36s, 36-48s, and 48-60s).
One segment (randomly chosen for each signal) is used for test, while the rest is used for training.
In the experiments, we adopt a subject-dependent setting as in \cite{jang2018eeg}, where the data of all subjects are used for training or test.

For each signal, we apply a three-second-long time window with two-second-long overlap to augment the dataset, while the window does not extend across a training segment and a test segment.
Hence, we obtain 44 to 46 training samples, and 10 test samples for each signal in a fold.
Note that the number of training samples differs depending on the position of the test segment in the fold, i.e., whether the test segment is in the middle of a signal or is the first/last segment.
As a result, we have 56,320 or 58,880 training samples and 12,800 test samples for the DEAP database.
In the case of the DREAMER database, the number of training samples is 18,216 or 19,044, and the number of test samples is 4,140.
% \color{black}

\subsubsection{Model details}
\label{sec:model_details}
\noindent
% \color{blue}
\textbf{Graph membership extraction.} $f_1$, $f_2$, and $f_3$ are two-layer fully-connected networks with exponential linear units (ELU) \cite{clevert2015fast} for activation.
Batch normalization \cite{ioffe2015batch} is used in their output layers.
$f_4$ consists of three fully-connected layers with ELU activation.
The number of neurons in each hidden layer and the intermediate output layer is 256 for DEAP and 512 for DREAMER, and the last layer has $K$ output neurons.
% \color{black}

\noindent
\textbf{Feature extraction.} Each dilated inception block consists of four 1-D convolutional layers having a kernel size of 3 and dilation rates of 1, 2, 4, and 8, respectively. Each layer has eight output channels, so $F=8 \times 4=32$.
Padding is used in the convolutional layers, so that the time length of the signal after each convolution is maintained.
Each dilated inception block is followed by ReLU activation, max-pooling with a size of $4$, and batch normalization.
The feature extraction module consists of three dilated inception blocks in a row. Hence, the time length of the output features is $T'=384/4^{3}=6$.

\noindent
% \color{blue}
\textbf{Graph neural network.} $g_1^k$ ($k=1,\ldots,K$), and $g_2$ are two-layer fully-connected networks having ReLU activation.
The number of neurons in each hidden layer and the output layer of $g_1^k$ is 256 for DEAP and 512 for DREAMER.
The output layer of $g_2$ uses the softmax function, whose size is equal to the number of classes (i.e., 40 for DEAP and 18 for DREAMER).
% \color{black}

% \color{blue}
\subsubsection{Implementation} Our model is trained with the Adam optimizer \cite{kingma2014adam} with $\beta_{1}=0.9$ and $\beta_{2}=0.999$ for 30 epochs to minimize the cross-entropy loss. The learning rate is set to $0.001$, and the batch size is set to 32.
We report the average accuracy along with the standard deviation value over the five folds.
Our model is implemented in PyTorch \cite{paszke2017automatic}, and the code is available at \url{https://github.com/ELEMKEP/bsc_lcs}.

With a single NVIDIA K80 GPU, the training procedure takes 10 to 15 minutes and 1.5 to 2 minutes for an epoch for the DEAP and the DREAMER databases, respectively.
For the inference, the model takes about 5ms for the DEAP database and about 2ms for the DREAMER database for a single sample.
When the effect of the number of training samples is excluded, the training time is approximately proportional to the number of electrodes.
This is because the operations in the proposed model such as membership extraction and feature extraction are applied for each electrode.
% \color{black}

\subsection{Performance comparison}

\subsubsection{Compared methods}

\noindent
\textbf{Non-GNN methods.} We adopt two types of traditional classification methods: k-nearest neighbor (k-NN), and support vector machine (SVM) using linear kernel \cite{zheng2015investigating}.
To obtain the input features from the EEG signals for these methods, we compute differential entropy (DE) for five frequency bands ($\delta$, $\theta$, $\alpha$, $\beta$, and $\gamma$), as presented in \cite{zheng2015investigating}.
The hyperparameters (k=15 for k-NN and the regularization parameter $C=0.1$ for SVM) are determined experimentally for the maximal performance of the methods.
% \color{blue}
In addition, we employ the RACNN method \cite{cui2020eeg} using a CNN that receives the raw signal as input in the form of 3-D tensor and performs 3-D convolution and 2-D convolution sequentially.
% \color{black}

\noindent
\textbf{GNN-based methods.} We use three baseline methods that utilize connectivity structures with GNNs.
First, the spatial-temporal graph convolutional network (ST-GCN) \cite{yan2018spatial} is employed, which performs graph convolutions in both the spatial domain and the temporal domain.
Here, we use the distance-based k-NN graph (k $=4$) for spatial domain graph convolution, while the temporal domain graph has edges between the same electrodes of consecutive time frames.
The raw signals are fed to the network as input.
Second, the method using a ChebNet \cite{jang2018eeg} is considered. The method uses a graph in which a given vertex (i.e., electrode) is expanded over eight frequency bands ($\delta$, $\theta$, $\alpha_{\text{low}}$, $\alpha_{\text{high}}$, $\beta_{\text{low}}$ $\beta_{\text{mid}}$, $\beta_{\text{high}}$, and $\gamma$) so that the number of vertices in the graph is eight times the number of electrodes.
A subgraph of a single frequency-band has the structure of a distance-based k-NN graph (k $=4$), and the vertices corresponding to the same electrode in different subgraphs are also connected.
As the vertex feature, the signal power of the band-pass filtered signal, which shows the best performance among various candidates in \cite{jang2018eeg}, is used.
The model consists of two graph convolutional layers and two fully-connected layers.
Third, the DGCNN method \cite{song2018eeg} is adopted.
The model uses a graph convolutional layer followed by a 1-D convolutional layer and a fully-connected layer.
We use the DE features for the five frequency bands ($\delta$, $\theta$, $\alpha$, $\beta$, and $\gamma$) for the input data and a distance-based k-NN graph (k $=4$) as the initial graph.
% \color{blue}
Fourth, the IAG method \cite{song2020instance} is used.
The DE features are used for input as in DGCNN, from which a graph structure is computed and processed by a GNN.
The resulting features of the electrodes in each region are aggregated, and a fully-connected network is applied for classification.
% \color{black}

\subsubsection{Results}

% \color{blue}
Table \ref{table:accuracy_experiment} summarizes the classification accuracy of the proposed model (using continuous softmax and three graph layers with a skip layer) and the existing methods.
It is clear that the proposed method yields significantly better performance than the other methods, especially for the DEAP database.
In particular, the superiority of our method to ST-GCN and ChebNet demonstrates that the multi-layer graph structure learned specifically towards classification is more effective than the manually determined ones.
In addition, the low performance of these GNN-based methods implies the difficulty of pre-defining an appropriate graph structure, which is not needed in the proposed method.
Compared to the methods that learn graph structures (i.e., DGCNN and IAG), the higher performance of our method supports that the feature extraction in our method works well.
Comparison between ChebNet and other two GNN-based methods (ST-GCN and DGCNN) is also notable.
ST-GCN and DGCNN use a simple electrode-to-vertex mapping to construct a single-layer graph having the number of vertices equal to the number of electrodes.
On the other hand, ChebNet uses an expanded graph from the multiple frequency bands, which gives more expressive power for the classifier and consequently yields better performance than ST-GCN and DGCNN.
IAG uses a similar approach to ChebNet, which constructs the expanded graph based on the frquency bands, so that IAG shows the performance comparable to ChebNet on the DREAMER database.
Quite surprisingly, k-NN shows good performance similarly to our method on DREAMER.
We found that many of the nearest neighbors for a test sample are from the signal of the same subject and the same stimulus.
% \color{black}

\begin{table}
  \centering
  \caption{\label{table:accuracy_experiment}Comparison of the classification accuracy.}
  \resizebox{0.99\columnwidth}{!}{
    \begin{tabular}{rrr}
      \toprule
                                        & DEAP            & DREAMER         \\
      \midrule
      k-NN                              & 11.2$\pm$0.82\% & 56.0$\pm$4.56\% \\
      SVM \cite{zheng2015investigating} & 3.55$\pm$2.01\% & 19.3$\pm$1.34\% \\
      RACNN \cite{cui2020eeg}           & 2.58$\pm$0.08\% & 32.0$\pm$3.94\% \\
      \midrule
      ST-GCN \cite{yan2018spatial}      & 7.07$\pm$0.40\% & 26.9$\pm$3.83\% \\
      ChebNet \cite{jang2018eeg}        & 10.3$\pm$1.38\% & 31.4$\pm$6.77\% \\
      DGCNN \cite{song2018eeg}          & 2.52$\pm$0.03\% & 19.3$\pm$2.10\% \\
      IAG \cite{song2020instance}       & 7.90$\pm$1.22\% & 31.8$\pm$4.53\% \\
      \midrule
      Proposed w/o graph learning       & 56.6$\pm$8.39\% & 48.1$\pm$7.97\% \\
      Proposed method                   & 73.5$\pm$8.07\% & 55.5$\pm$7.59\% \\
      \bottomrule
    \end{tabular}}
\end{table}

\subsection{Ablation studies}

\subsubsection{Module ablation}
We design an ablation setting to examine the importance of graph structure learning.
For this, we test our model without graph structure extraction including the graph membership extraction and graph sampling modules so that only the lower part in Figure \ref{fig:overall} is used for training and inference.
Instead, an unweighted complete graph is used in the GNN part, so that the classification depends only on the extracted features.

The results of this setting are shown in Table \ref{table:accuracy_experiment} as ``Proposed w/o graph learning'', which show deteriorated performance.
This demonstrates the importance of the learned graph structure and the synergy effect between graph learning and feature extraction in the proposed model.
In addition, this implies that connecting non-related vertices significantly harms the classification performance.

\begin{table}
  \centering
  \caption{\label{table:accuracy_graph}Accuracy of the proposed method for various combinations of the number of graph layers, the graph sampling method, and the existence of the skip layer. The top cases for each graph sampling method are highlighted.}
  \resizebox{0.99\columnwidth}{!}{
    \begin{tabular}{clccc}
      \toprule
      \multicolumn{1}{c}{Database} & \multicolumn{1}{c}{\#Layers} &
      \multicolumn{1}{c}{STO}      &
      \multicolumn{1}{c}{DET}      &
      \multicolumn{1}{c}{CON}                                                                                                                      \\
      \midrule
      \multirow{5}{*}{DEAP}        &
      1+skip                       & \textbf{72.6$\pm$5.84\%}     & 74.2$\pm$3.82\%          & 70.0$\pm$9.28\%                                     \\
                                   & 2                            & 65.9$\pm$7.03\%          & \textbf{75.4$\pm$8.42\%} & 64.2$\pm$4.19\%          \\
                                   & 2+skip                       & 64.6$\pm$8.79\%          & 74.0$\pm$4.71\%          & 68.6$\pm$6.20\%          \\
                                   & 3                            & 61.3$\pm$6.56\%          & 74.3$\pm$3.59\%          & 61.3$\pm$6.30\%          \\
                                   & 3+skip                       & 72.1$\pm$7.60\%          & 73.6$\pm$4.41\%          & \textbf{73.5$\pm$8.07\%} \\
      \midrule
      \multirow{5}{*}{DREAMER}     &
      1+skip                       & 51.9$\pm$10.3\%              & 50.9$\pm$11.1\%          & 54.6$\pm$10.0\%                                     \\
                                   & 2                            & 52.0$\pm$8.97\%          & 51.9$\pm$6.33\%          & 52.2$\pm$12.6\%          \\
                                   & 2+skip                       & 52.1$\pm$8.24\%          & \textbf{55.1$\pm$9.05\%} & 51.3$\pm$11.7\%          \\
                                   & 3                            & \textbf{52.5$\pm$10.3\%} & 53.5$\pm$6.29\%          & 54.2$\pm$12.6\%          \\
                                   & 3+skip                       & 51.0$\pm$9.63\%          & 50.7$\pm$10.6\%          & \textbf{55.5$\pm$7.59\%} \\
      \bottomrule
    \end{tabular}}
\end{table}

\subsubsection{Hyperparameters for graph extraction}
In Table \ref{table:accuracy_graph}, we show the accuracy of our model for various combinations of the graph sampling method, the number of graph layers ($K$), and the existence of the skip layer.
% \color{blue}
For the DEAP database, the results show that the sampling method is an important hyperparameter, while the number of graph layers including the existence of the skip layer is relatively minor.
The deterministic thresholding method shows the best performance overall.
The main difference of deterministic thresholding from the other sampling methods is that it can assign explicit connectivity in multiple graph layers, which seems to lead to slight performance gains.
In the case of the DREAMER database, the performance variation across the different conditions is relatively low, probably due to the smaller number of electrodes and consequently the smaller size of graphs.
% For the DREAMER database, the hyperparameters show relatively minor effects comparing to the DEAP database.
% The continuous sampling method shows larger standard deviation than stochastic sampling or the deterministic thresholding, in terms of the time-domain folds.
% \color{black}

\subsection{Graph consistency analysis}
\label{sec4:experiments_consistency}

Since no ground truth for the graph structure is available, it is not possible to evaluate the correctness of the extracted graphs.
As an alternative way to evaluate the quality of the extracted graphs, we consider their consistency over repeated experiments. To measure the consistency, we calculate the dissimilarity between the extracted graph structures for all pairwise combinations of repetitions.
A low level of dissimilarity means a high level of consistency, which indicates that the extracted graph structures are reliable and meaningful for the classification.
One issue in measuring the dissimilarity between two graphs is the permutation ambiguity of graph layers, because we do not impose any restriction on the order of the graph layers during training.
To deal with this, we perform an exhaustive search to find the best matching permutation of graph layers yielding the smallest dissimilarity measure.

\begin{algorithm}[ht]
  \caption{Computing graph dissimilarity}\label{alg:consistency}
  \textbf{Input:} $Diss(\cdot,\cdot), W_{(n)}\in\mathbb{R}^{K \times |E|}, W_{(m)}\in \mathbb{R}^{K \times |E|}$ \\
  \textbf{Output:} $D_{nm}^*$ %, P^*$
  \begin{algorithmic}[1]
    \State $M = \text{GetPerm}(1,...,K)$ \Comment{Make a set of permutations}
    \State $D_{nm}^* \gets \infty$ \Comment{Initial value}
    \For {$P$ in $M$} \Comment{Pick a permutation of $(1,...,K)$}
    \State $W_{(m)}^{'} \gets \text{Permute } (W_{(m)}, P)$
    \State $D=Diss\left(W_{(n)}, W_{(m)}^{'}\right)$
    \If {$D < D_{nm}^*$}
    \State $D_{nm}^* \gets D$
    \EndIf
    \EndFor
  \end{algorithmic}
\end{algorithm}

The overall procedure to compute dissimilarity is summarized in Algorithm \ref{alg:consistency}.
Here, $Diss$ computes the dissimilarity between the adjacency matrices of two graph layers, for which we adopt the mean of absolute differences between the two matrices.
Note that we do not have the issue of the graph isomorphism because the order of vertices is clearly identified as distinguished EEG electrodes.
$W_{(n)}$ and $W_{(m)}$ correspond to the extracted graph structures from a pair of repetitions.
$M$ is the set of all possible permutations of graph layers.
The dissimilarity value ($D$) between $W_{(n)}$ and each $W^{'}_{(m)}$, which corresponds to a permutation of $W_{(m)}$ in $M$, is calculated.
The smallest dissimilarity value among all permutations, $D_{nm}^{*}$, is obtained as the final dissimilarity between $W_{(n)}$ and $W_{(m)}$.
After the dissimilarities for all pairs of repetitions are computed, we obtain the final consistency score by subtracting the average dissimilarity value from 1:
\begin{equation}
  \label{eq:consistency_score}
  \phantom{,} \mathrm{Consistency\_score}=1 - \frac{2}{R(R-1)}\ \sum_{n=1}^{R-1} \sum_{m=n+1}^{R} D^*_{nm} ,
\end{equation}
where $R$ is the number of repetitions ($R=5$ in our case).

% \color{blue}
Table \ref{table:distance} shows the results of consistency analysis in percentage.
Overall, high consistency levels (about 60 to 100\%) are observed across different conditions, indicating that the extracted graphs contain meaningful representations of the data.
As in Table \ref{table:accuracy_graph}, the main factor affecting the graph consistency is the sampling method.
In the DEAP database, the deterministic thresholding method shows higher consistency than the other two methods.
The highest consistency of the case of ``1+skip'' is from excessive edge activations, where a large number of edges are activated for all repeated experiments.
This seems to be due to insufficient expressive power in the single-layer graph setting.
In the case of the DREAMER database, the overall tendency is similar to the DEAP database, while the differences between the sampling methods are smaller.
% \color{black}

% Version 2.0
\begin{table}
  \centering
  \caption{\label{table:distance}Consistency scores of the extracted graphs for different conditions.}
  \resizebox{0.99\columnwidth}{!}{
    \begin{tabular}{clccc}
      \toprule
      \multicolumn{1}{c}{Database} & \multicolumn{1}{c}{\#Layers} &
      \multicolumn{1}{c}{STO}      &
      \multicolumn{1}{c}{DET}      &
      \multicolumn{1}{c}{CON}                                                                \\
      \midrule
      \multirow{5}{*}{DEAP}        &
      1+skip                       & 99.9\%                       & 99.4\% & 99.9\%          \\
                                   & 2                            & 74.2\% & 82.8\% & 78.6\% \\
                                   & 2+skip                       & 60.9\% & 96.8\% & 59.8\% \\
                                   & 3                            & 72.3\% & 87.6\% & 80.4\% \\
                                   & 3+skip                       & 74.3\% & 89.3\% & 76.7\% \\
      \midrule
      \multirow{5}{*}{DREAMER}     &                                                         % TODO
      1+skip                       & 62.4\%                       & 64.7\% & 66.5\%          \\
                                   & 2                            & 68.6\% & 76.2\% & 63.1\% \\
                                   & 2+skip                       & 65.2\% & 69.6\% & 64.4\% \\
                                   & 3                            & 67.5\% & 73.0\% & 69.2\% \\
                                   & 3+skip                       & 72.8\% & 73.1\% & 73.9\% \\
      \bottomrule
    \end{tabular}}
\end{table}

\begin{figure*}[!ht]
  \small
  \centering
  \resizebox{0.95\textwidth}{!}{
    \begin{tabular}{c}
      % Video 19: high valence
      % Video 30: low valence
      \begin{tabular}{>{\centering\arraybackslash}m{4cm}>{\centering\arraybackslash}m{4cm}>{\centering\arraybackslash}m{4cm}}
        \includegraphics[trim={10 30 10 0},clip,width=4cm]{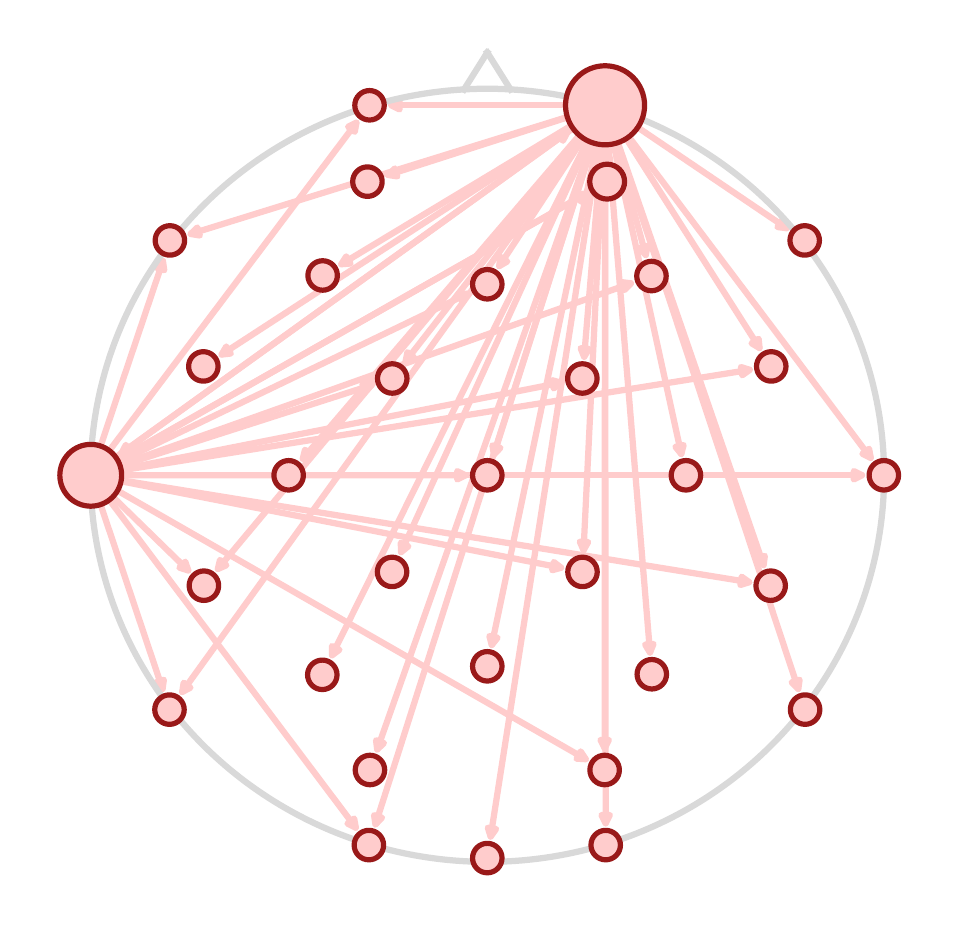} &
        \includegraphics[trim={10 30 10 0},clip,width=4cm]{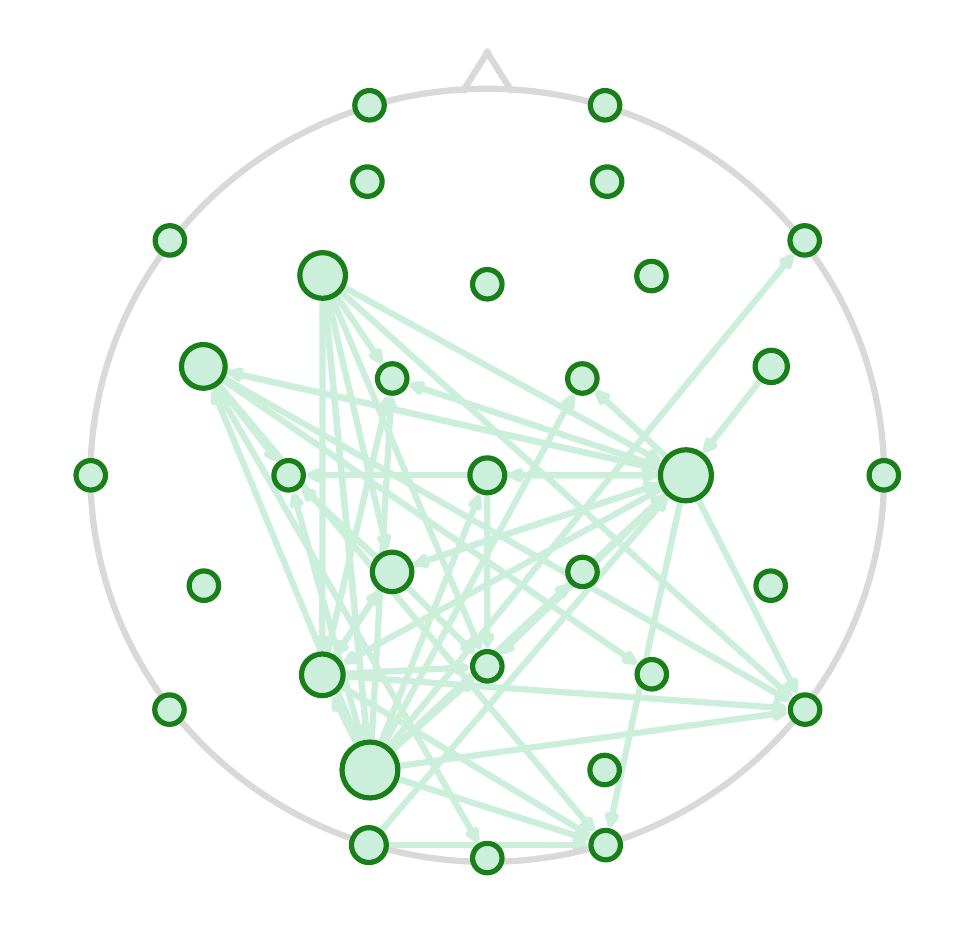} &
        \includegraphics[trim={10 30 10 0},clip,width=4cm]{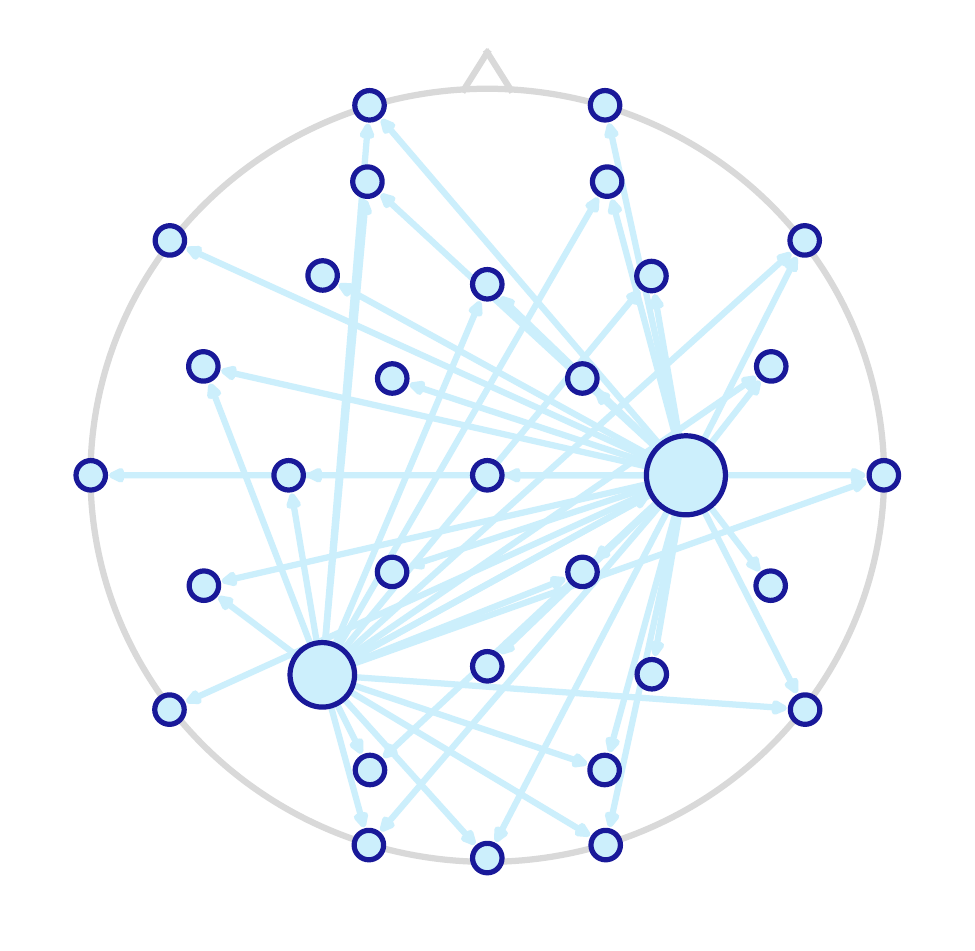}               \\
        (a1)                                                                                 & (a2) & (a3)
      \end{tabular} \\
      \begin{tabular}{>{\centering\arraybackslash}m{4cm}>{\centering\arraybackslash}m{4cm}>{\centering\arraybackslash}m{4cm}}
        \includegraphics[trim={10 30 10 0},clip,width=4cm]{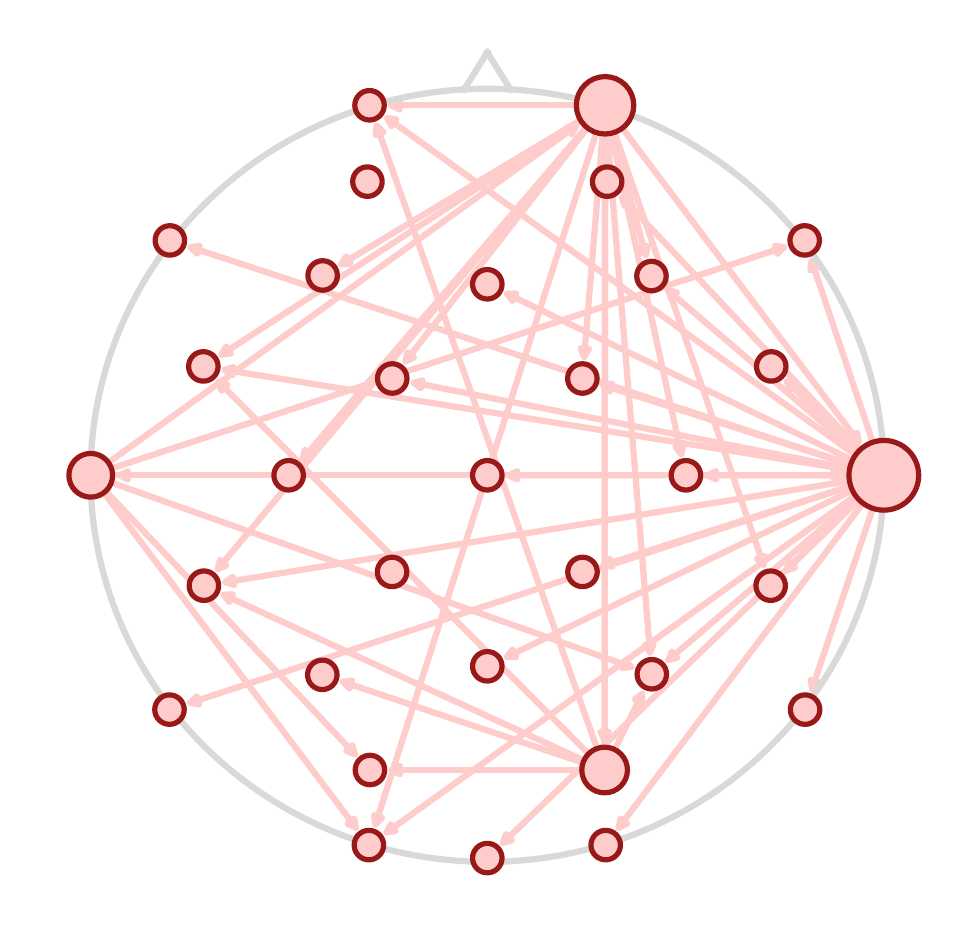} &
        \includegraphics[trim={10 30 10 0},clip,width=4cm]{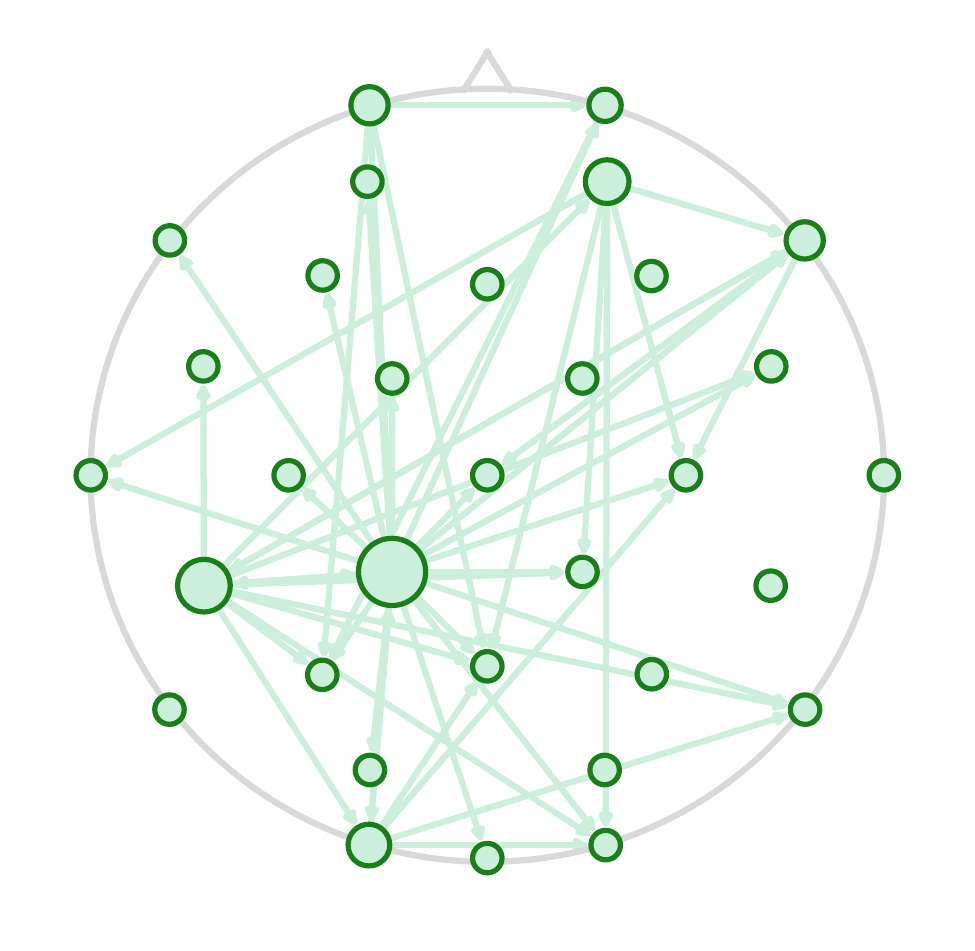} &
        \includegraphics[trim={10 30 10 0},clip,width=4cm]{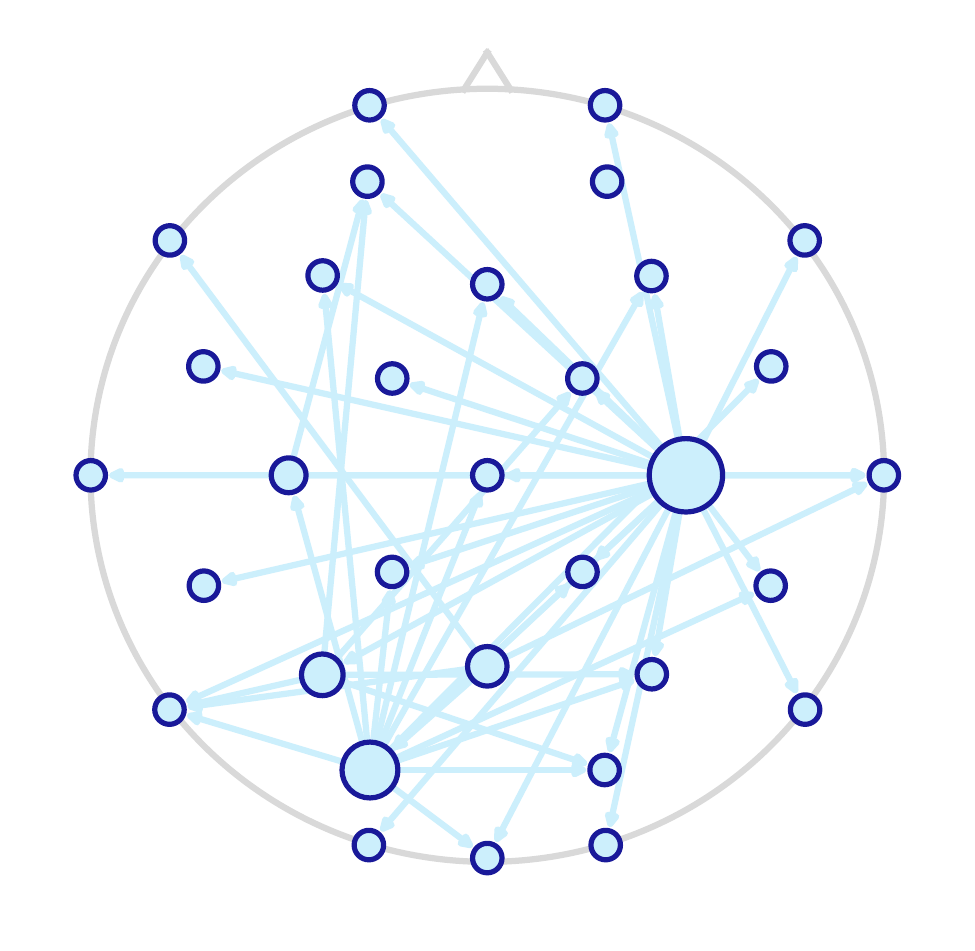}               \\
        (b1)                                                                                 & (b2) & (b3)
      \end{tabular} \\
      \begin{tabular}{>{\centering\arraybackslash}m{4cm}>{\centering\arraybackslash}m{4cm}>{\centering\arraybackslash}m{4cm}}
        \includegraphics[trim={10 30 10 0},clip,width=4cm]{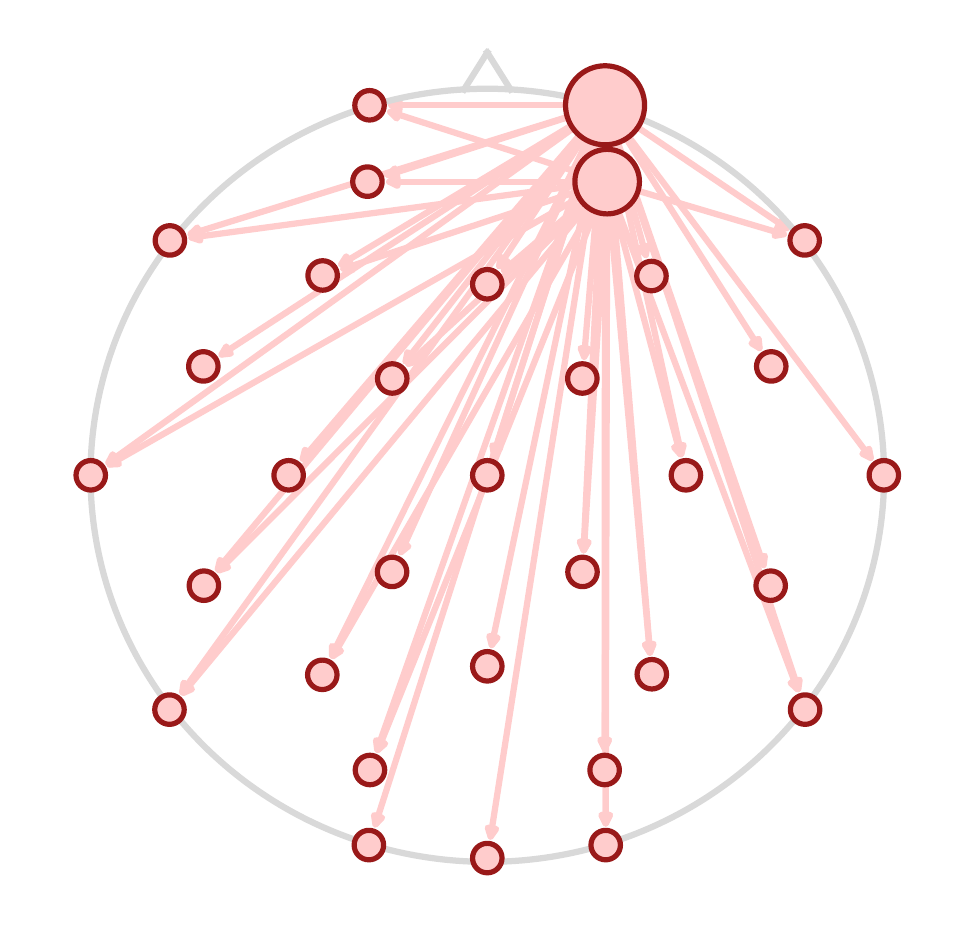} &
        \includegraphics[trim={10 30 10 0},clip,width=4cm]{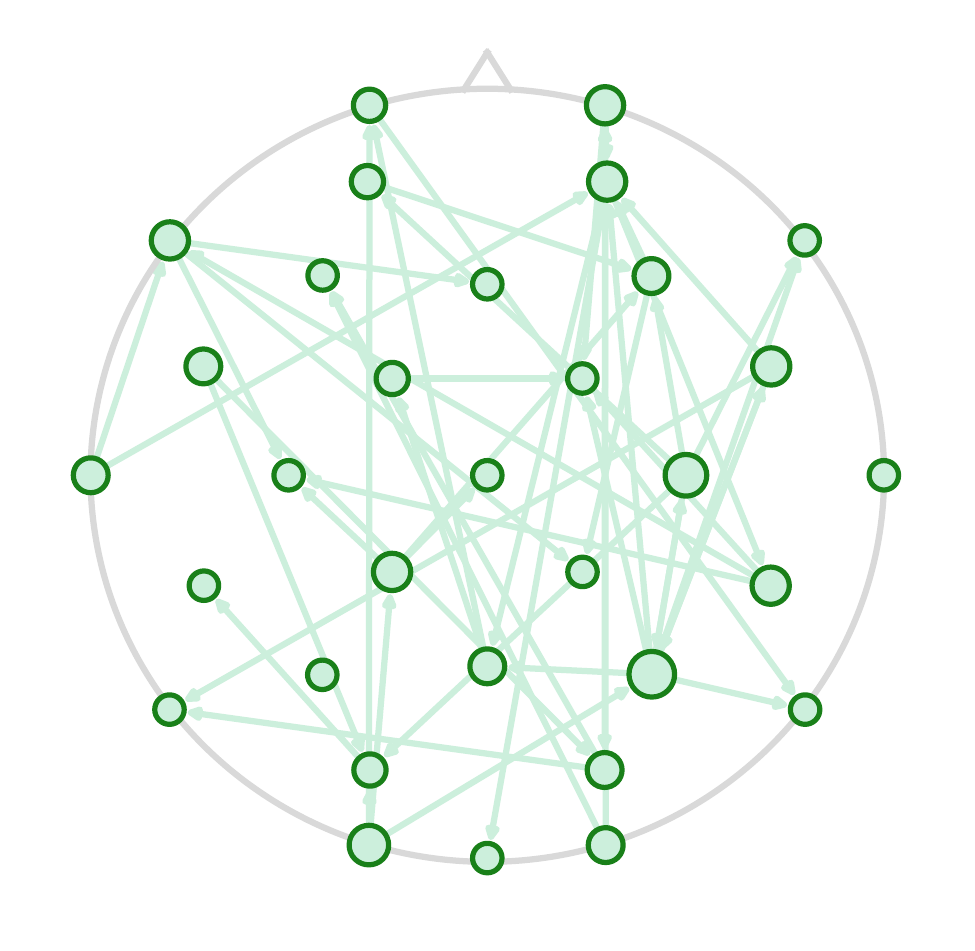} &
        \includegraphics[trim={10 30 10 0},clip,width=4cm]{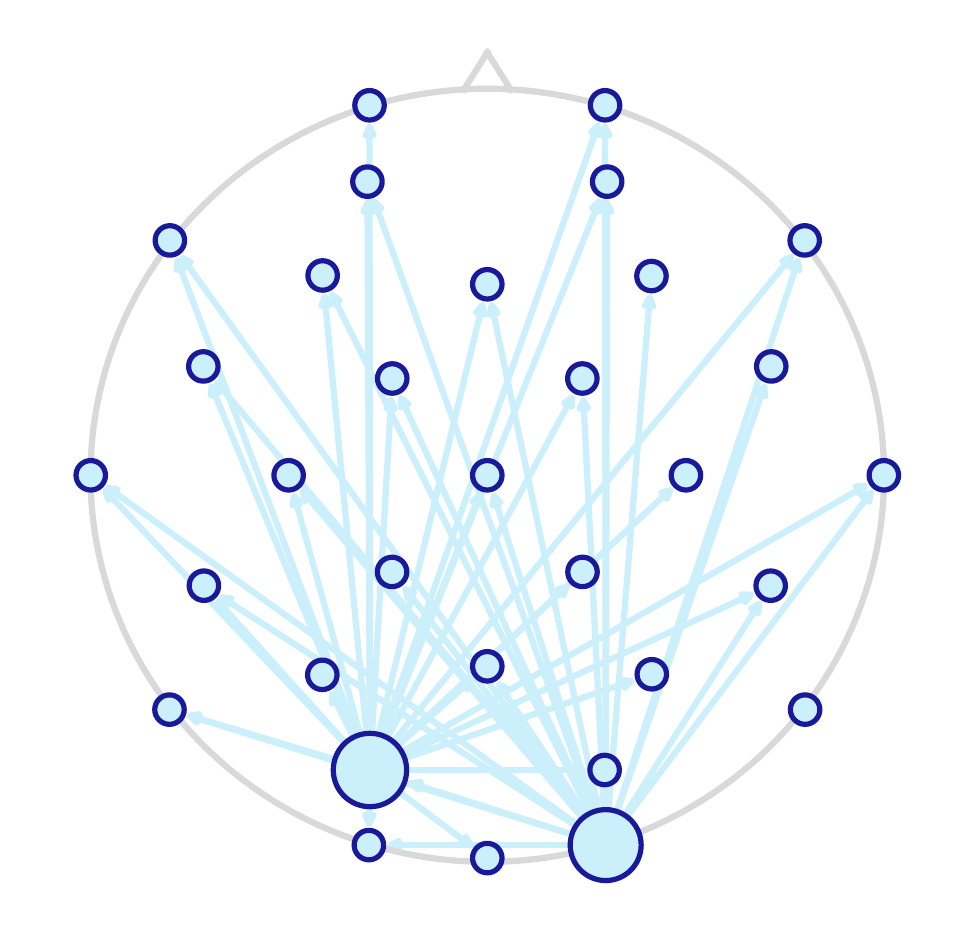}               \\
        (c1)                                                                                & (c2) & (c3)
      \end{tabular} \\
    \end{tabular}}
  \caption{\label{fig:graphs_deap} Representative graph structures from the DEAP database for (a1-3) all test dataset, (b1-3) the video stimulus having positive valence, and (c1-3) the video stimulus having negative valence. Each graph (red (1), green (2), and blue (3)) corresponds to each graph layer. The size of a vertex is proportional to its out-degree. The in-degree is not explicitly shown because it is similar across all vertices.}
\end{figure*}

\subsection{Graph structure analysis}

We examine the obtained graph structures in order to understand the learned representations from the perspective of affective cognitive responses.
Since the graph structure is different for each data and each repeated experiment, we obtain a representative graph structure by aggregating multiple graph structures.
We define the representative graph structure as the one that has edges corresponding to the most frequently appearing edges (top-5\%) in all data.
Here, we consider only the DEAP database, due to its rich coverage of the scalp with a sufficiently large number of electrodes compared to the DREAMER database.

Figure \ref{fig:graphs_deap} shows the obtained representative graph structures.
In Figures \ref{fig:graphs_deap}(a1), \ref{fig:graphs_deap}(a2), and \ref{fig:graphs_deap}(a3), the ``default graphs'' that contain edges activated most frequently regardless of the task (i.e., across different video stimuli) are shown by obtaining the representative graph for all test data.
Figures \ref{fig:graphs_deap}(b1), \ref{fig:graphs_deap}(b2), and \ref{fig:graphs_deap}(b3) correspond to the representative graph structures for the video stimuli corresponding to the positive valence rating, while Figures \ref{fig:graphs_deap}(c1), \ref{fig:graphs_deap}(c2), and \ref{fig:graphs_deap}(c3) show the graph structures for the video with the negative valence ratings.

% \color{blue}
Figures \ref{fig:graphs_deap}(a1), \ref{fig:graphs_deap}(b1), and \ref{fig:graphs_deap}(c1) show strong activations at the right frontal lobe.
It is consistently reported that the asymmetry of the left and right frontal lobes is associated with the valence of emotion \cite{harmon-jones10role,reznik18frontal}.
In Figures \ref{fig:graphs_deap}(a1) and \ref{fig:graphs_deap}(b1), the activation from the right frontal lobe to the temporal lobes is observed.
Especially, Figure \ref{fig:graphs_deap}(b1), which corresponds to the stimulus with positive valence, shows stronger activation of the right temporal lobe than Figure \ref{fig:graphs_deap}(a1) (the default graph).
In contrast, Figure \ref{fig:graphs_deap}(c1), which is for the stimulus having negative valence, shows weak activation at the temporal lobes while it shows stronger activation at the right frontal area than Figures \ref{fig:graphs_deap}(a1) and (b1).
It is known that the temporal lobe is largely related to the processing of complex visual stimuli such as scenes \cite{alarcao17emotions}.
Thus, we can say that the first graph layer represents the mental state under exposure to emotional visual stimuli.

The second graph layers (Figures \ref{fig:graphs_deap}(a2), \ref{fig:graphs_deap}(b2), and \ref{fig:graphs_deap}(c2)) show activations over a wide region covering the fronto-central area and parietal area, which is relatively less emphasized in the other layers.
It is known that the fronto-central area is related to the motor-related perception \cite{hauk04neuro}, and the parietal area is related to the visual tracking \cite{battelli2009role}.
Therefore, it is thought that the second layer accounts for the perception of the motion information appearing in the video stimuli.

In Figures \ref{fig:graphs_deap}(b3) and (c3), the differences between the activations in the occipital lobe and right central area are noticeable.
Especially, the activation of the occipital lobe is characteristic compared to the other two graph layers.
It is reported that the activation of the occipital lobe is changed by the emotional factors of the stimuli \cite{gerdes2010brain}.
In addition, the activation of the occipital lobe is also related to the visual motion \cite{grill-spector04human,kamps16occipital}, which explains the content difference between the two stimuli.
Therefore, it seems that the third graph layer supports the other two layers by accounting for emotional and visual factors.
% \color{black}

From this qualitative analysis, we can conclude that the proposed method can successfully model structural patterns of the EEG signal with the multi-layer graph structure.

\section{Conclusion}
\label{sec5:conclusion}

We have proposed an end-to-end deep network that can extract an appropriate directed multi-layer graph structure from the given raw EEG signals for classification without any a priori information about the desirable structure.
The experimental results showed that this data-driven approach for learning the graph structure significantly improves the classification performance in comparison to the other GNN approaches based on manually defined features and graphs.
It was also shown that the extracted graph structures are reliable, and consistent with the known brain activation patterns of cognitive responses to emotional visual stimuli.

In the proposed approach, the computational burden is largely determined by the number of EEG electrodes.
Since the extracted membership and graph structures have the dimension of the square of the number of electrodes, the model is memory-demanding if a large number of electrodes is used as in the SEED\cite{zheng2015investigating} and MPED\cite{song2019mped} databases.
Thus, designing memory-efficient modules and reducing the dependency on the number of electrodes will be desirable topics of future work for a wider application of the proposed end-to-end approach to connectivity inference-based classification.

Another topic for future work is the subjectivity of EEG.
Human's emotional experience is subjective, i.e., different people may respond differently to the same stimuli \cite{koelstra2011deap}.
In addition, each subject has her/his unique EEG pattern, which leads to the difficulty of extracting task-relevant EEG patterns from recorded signals \cite{hu2019ten}.
The subjectivity of emotional responses and EEG patterns becomes an obstacle to obtaining high performance of EEG classification over multiple subjects.
Several approaches have been proposed, e.g., disentangling subjective features via learning \cite{moon2020madenet} or applying transfer learning \cite{zheng2016personalizing}.
However, no such work has been done for EEG connectivity.
In this regard, developing methods to separate subjective and task-relevant connectivity will be an impactful direction for future work.

\ifCLASSOPTIONcompsoc
  \section*{Acknowledgments}
\else
  \section*{Acknowledgment}
\fi

This work was supported by the Artificial Intelligence Graduate School Program (Yonsei University, 2020-0-01361).

\ifCLASSOPTIONcaptionsoff
  \newpage
\fi



\bibliographystyle{IEEEtran}
\bibliography{IEEEabrv, ./refs.bib}


% \begin{IEEEbiography}[{\includegraphics[width=1in,height=1.25in,clip,keepaspectratio]{mshell}}]{Michael Shell}

\begin{IEEEbiography}[{\includegraphics[width=1in,height=1.25in,clip]{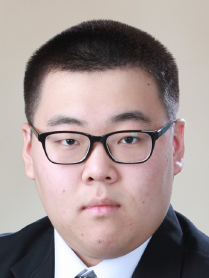}}]{Soobeom Jang}
  received the B.S. degree from Yonsei University, Seoul, South Korea, in 2014, where he is currently working toward the Ph.D. degree.
  His research interests include graph neural network, deep learning for multimedia applications, and social multimedia analysis.
\end{IEEEbiography}
% \vskip 0pt plus -1fil

\begin{IEEEbiography}[{\includegraphics[width=1in,height=1.25in,clip]{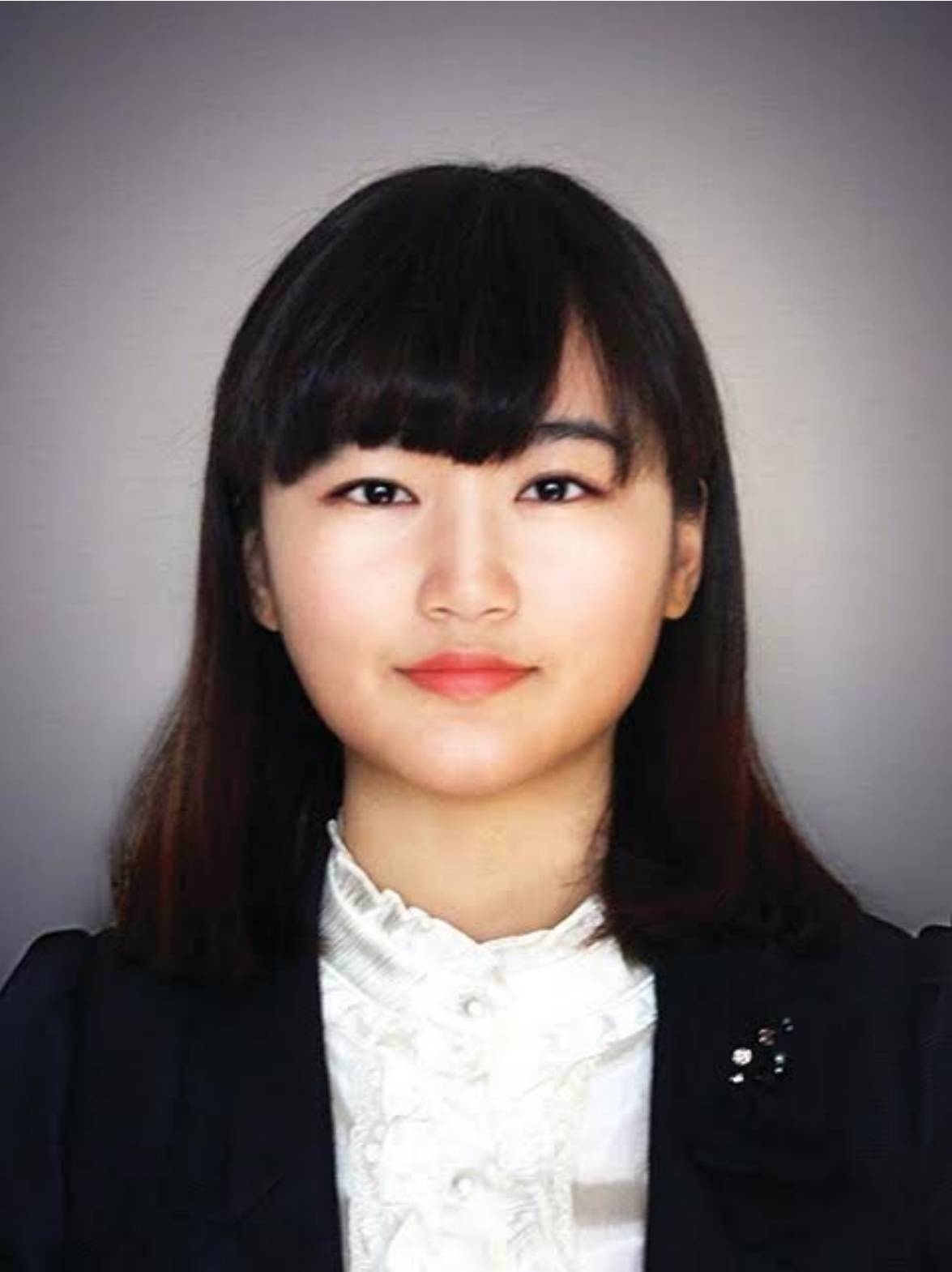}}]{Seong-Eun Moon}
  received the B.S. degree in mechanical engineering from Chiba University, Chiba, Japan, in 2013.
  She received the Ph.D. degree from School of Integrated Technology, Yonsei University, Seoul, South Korea, in 2020.
  Her research interests include multimedia signal processing, physiological signal processing, and machine learning.
\end{IEEEbiography}
% \vskip 0pt plus -1fil

\begin{IEEEbiography}[{\includegraphics[width=1in,height=1.25in,clip]{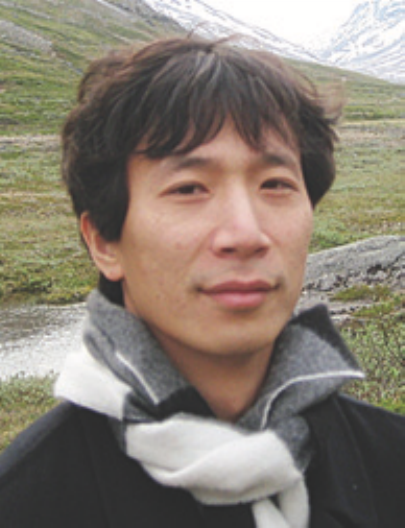}}]{Jong-Seok Lee}
  received the Ph.D. degree in electrical engineering and computer science from Korea Advanced Institute of Science and Technology, Korea. He worked as a Researcher in the Swiss Federal Institute of Technology in Lausanne (EPFL), Switzerland. Currently, he is a Professor in the School of Integrated Technology, Yonsei University, Korea. His research interests include multimedia signal processing and machine learning. He is an author or co-author of more than 200 publications. He serves as an editor of the IEEE Communications Magazine and Signal Processing: Image Communication. He is a senior member of the IEEE.
\end{IEEEbiography}


% % insert where needed to balance the two columns on the last page with
% %\newpage



%\vfill


\end{document}